\documentclass[letterpaper, 10 pt, journal, twoside]{ieeetran}
\usepackage{graphicx} 
\usepackage{cite}
\usepackage{algorithm} 
\usepackage{algorithmic} 
\usepackage{amsmath, amssymb} 
\usepackage{booktabs} 
\usepackage{url}
\usepackage{hyperref}
\usepackage[font=small]{caption}
\usepackage{colortbl}  
\usepackage{xcolor}    
\newcolumntype{g}{>{\columncolor{gray!20}}c}

\title{A Backbone for Long-Horizon Robot Task Understanding}

\author{Xiaoshuai Chen$^{1}$, Wei Chen$^{1}$, Dongmyoung Lee$^{1}$, Yukun Ge $^{1}$, Nicolas Rojas $^{2}$, and Petar Kormushev $^{1}$%

\thanks{Manuscript received: August, 14, 2024; 
November, 17, 2024; December, 18, 2024.}

\thanks{This paper was recommended for publication by Editor Vincze, Markus upon evaluation of the Associate Editor and Reviewers' comments. \textit{(Corresponding author: Xiaoshuai Chen.)}} 

\thanks{$^{1}$Xiaoshuai Chen, Wei Chen, Dongmyoung Lee, Yukun Ge, and Petar Kormushev are with the Dyson School of Design Engineering, Imperial College London, UK
        {\tt\footnotesize cx119@ic.ac.uk; w.chen21@imperial.ac.uk; d.lee20@imperial.ac.uk; yukun.ge20@imperial.ac.uk; p.kormushev@imperial.ac.uk}}%
\thanks{$^{2} $Nicolas Rojas is with The AI Institute, Cambridge, MA, USA.
        {\tt\footnotesize nrojas@ieee.org}}%
\thanks{Digital Object Identifier (DOI): see top of this page.}
}

\begin{document}
\maketitle

\begin{abstract}

End-to-end robot learning, particularly for long-horizon tasks, often results in unpredictable outcomes and poor generalization. 
To address these challenges, we propose a novel \textit{Therblig-Based Backbone Framework (TBBF)} as a fundamental structure to enhance interpretability, data efficiency, and generalization in robotic systems. TBBF utilizes expert demonstrations to enable therblig-level task decomposition, facilitate efficient action-object mapping, and generate adaptive trajectories for new scenarios. The approach consists of two stages: offline training and online testing. During the offline training stage, we developed the \textit{Meta-RGate SynerFusion (MGSF)} network for accurate therblig segmentation across various tasks. In the online testing stage, after a one-shot demonstration of a new task is collected, our \textit{MGSF} network extracts high-level knowledge, which is then encoded into the image using \textit{Action Registration (ActionREG)}. Additionally, \textit{Large Language Model (LLM)-Alignment Policy for Visual Correction (LAP-VC)} is employed to ensure precise action registration, facilitating trajectory transfer in novel robot scenarios. Experimental results validate these methods, achieving 94.37\% recall in therblig segmentation and success rates of 94.4\% and 80\% in real-world online robot testing for simple and complex scenarios, respectively. Supplementary material is available at: 
\href{https://sites.google.com/view/therbligsbasedbackbone/home}{https://sites.google.com/view/therbligsbasedbackbone/home}

\begin{IEEEkeywords}
Deep Learning in Grasping and Manipulation; Manipulation Planning; Learning from Demonstration
\end{IEEEkeywords}

\end{abstract}

\section{Introduction}

\IEEEPARstart{U}{nderstanding} robot tasks encompasses several key stages: sensing the environment, recognizing task-related objects, making decisions, and planning trajectories. Recently, data-driven methods, especially deep learning algorithms, have greatly advanced the field of robotics. While deep learning excels in object recognition and reinforcement learning aids in trajectory planning, these models often struggle to generalize beyond trained scenarios, especially in long-horizon tasks. Thus, improving generalization is crucial for adapting to diverse, dynamic, real-world situations effectively.

Traditional approaches to robot learning typically require large datasets and focus on simpler tasks like pick-and-place operations \cite{wu2023unleashing}. These end-to-end methods have shown effectiveness in mapping simple actions to objects within controlled environments \cite{jing2023exploring}. However, when it comes to long-horizon tasks and cluttered scenarios involving multiple steps, multiple unseen objects and intricate object interactions—such as liquid pouring—their effectiveness diminishes. In these situations, actions-objects mapping becomes significantly more challenging. End-to-end systems become inefficient and difficult to train, requiring vast amounts of data and computational resources \cite{sharma2023self}. Moreover, models trained on specific scenarios often lack the flexibility to adapt to new environments or task variations.

To address these limitations, we propose a novel framework that efficiently reconstructs long-horizon robot tasks by extracting action backbone and registering context information (see Fig. \ref{Fig1_intro}). Inspired by the concept of therbligs \cite{therbig}, a set of elemental motions, we introduce the fundamental architecture Therblig-Based Backbone Framework (TBBF). It systematically decomposes complex tasks into fundamental action units—termed "therbligs." These therbligs form the foundation for detailed task segmentation, action registration, task-related object reasoning, and trajectory adaptation to new layouts. Thus, our framework offers an interpretable, efficient, and scalable solution for robot task understanding, providing several key benefits:

\begin{figure}[t]
\centering
\includegraphics[width=\linewidth]{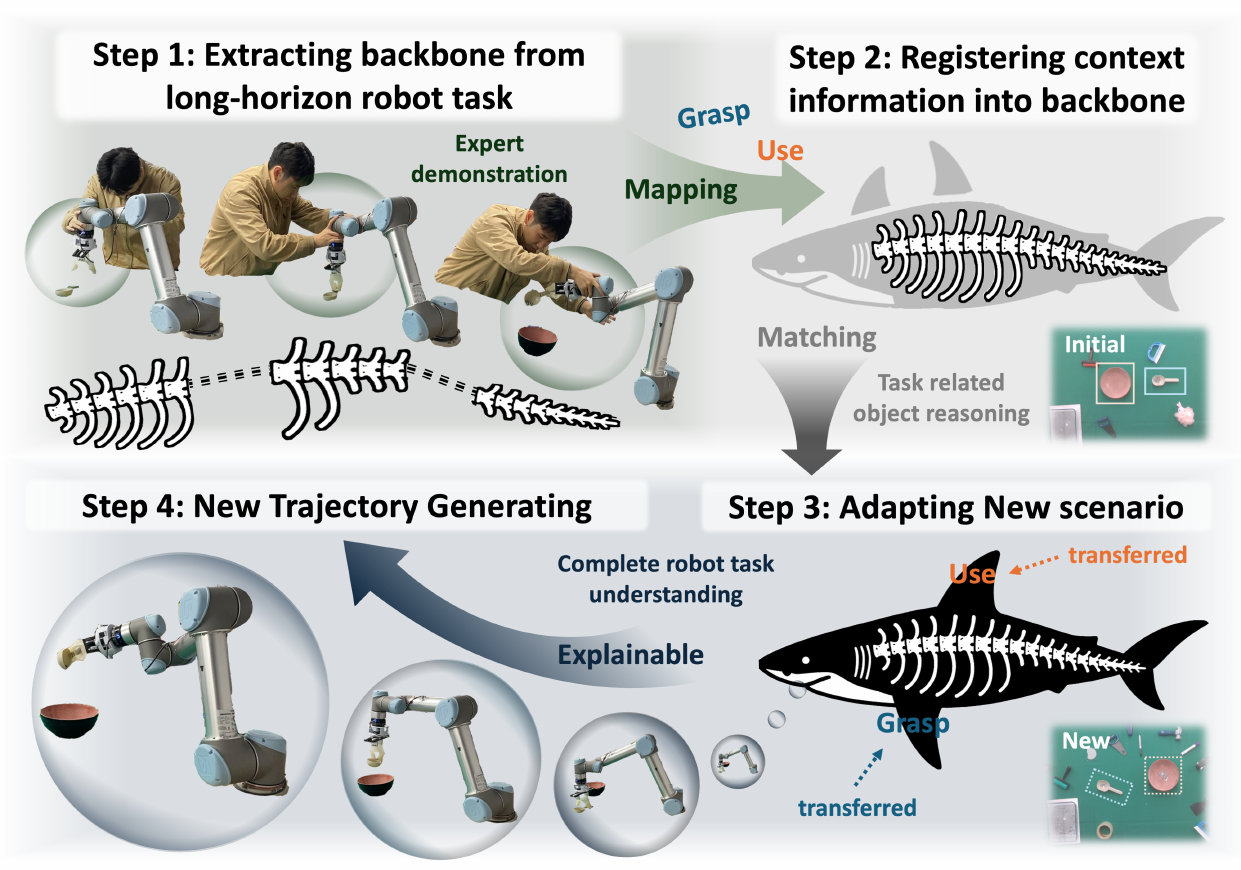}
\caption{Concept of the Proposed Robot Task Understanding System: extracts key backbone of complex tasks and uses context from a single demonstration to understand relevant objects and actions. }
\label{Fig1_intro}
\end{figure}

\renewcommand{\arraystretch}{1.3}  

\begin{table*}[ht]
    \centering
    \vspace{5pt} 
    \caption{Comparison of Capacities of Different Robot Systems}
    \vspace{-3pt}
    \label{tab:system_compare}
    \Large
    \resizebox{\textwidth}{!}{  
    \begin{tabular}{c g ccccccc}
        \specialrule{2pt}{1pt}{1pt} 
        \textbf{Capacity / System} & \textbf{TBBF(Ours)} & \textbf{Q-attention \cite{q-attention}} & \textbf{PerAct \cite{peract}} & \textbf{Diffuser \cite{diffuser}} & \textbf{MimicPlay \cite{wang2023mimicplay}} & \textbf{GLiDE \cite{llm1wang2024grounding}} & \textbf{Coarse to Fine \cite{coarse2fine}} \\
        \specialrule{2pt}{1pt}{1pt} 

        Data Efficiency & \textbf{High} & Moderate & Moderate & Low & High & High & Moderate\\
    
        Task Horizon & \textbf{Long-Horizon} & Short-Horizon & Mixed-Horizon  & Short-Horizon  & Long-horizon & Short-Horizon & Short-Horizon \\
        
        Task Interpretability & \textbf{Well-Structured} & Unstructured & Partially Structured  & Unstructured  & Partially Structured & Well-Structured & Partially Structured \\
        
        Task Diversity & \textbf{Wide Domain} & Narrow Domain & Wide Domain  & Narrow Domain & Wide Domain & Narrow Domain & Wide Domain \\
        
        Task Generalization & \textbf{One-shot} & Few-shot & 53 shots & 10k shots  & 20\&40 shots & Few-shot & One-shot \\
        
        Pre-train Scenarios & \textbf{Non-essential} & Required & Required  & Required  & Required & Required & Required \\
        
        Scenario Complexity & \textbf{Complex} & Moderate & Moderate  & Moderate  & Moderate & Simple & Simple \\

        Multi-modal Fusion & \textbf{Action, Vision, LLM} & Vision & Vision, LLM  & Action, Vision  & Action, Vision & Action, Vision, LLM & Action, Vision \\
        \specialrule{2pt}{1pt}{1pt} 
    \end{tabular}
    }
    \noindent\begin{minipage}{\textwidth}
    \scriptsize
    \vspace{3pt}
    * Data efficiency evaluates the amount and type of input data required by the system to learn and perform tasks. The Task Horizon indicates whether the system includes long-horizon tasks. Long-horizon tasks require extended sequences of actions, typically involving more than 10 individual steps \cite{pirk2020modeling}. Mixed-Horizon refers to tasks that contain both short and long horizons. Task Explanation assesses the clarity and interpretability of the system's decision-making process, such as action, context information and trajectory transformation. Task diversity assesses type of operation, object diversity, and environmental constraints. Scenario Complexity evaluates the number, diversity, arrangement of objects, and system's ability to adapt to new layouts (more details in our project website additional materials: \href{https://sites.google.com/view/therbligsbasedbackbone/home}{https://sites.google.com/view/therbligsbasedbackbone/home}).
    \end{minipage}
    
\end{table*}

Interpretability: TBBF enables transparency in the robot learning and execution processes by structuring tasks into distinct therblig sequences. This breakdown supports explainability by allowing each action in a complex task to be represented as interpretable units, making it possible to monitor, analyze, and generalize each action segment clearly and accurately. 

Data Efficiency and Robustness: By focusing only on task-relevant actions and objects, TBBF allows the system to disregard irrelevant objects in cluttered scenarios, improving robustness and reducing the need for re-training. This selective attention to critical elements also enables our system to learn from one-shot demonstrations, making it far more data-efficient than models that require extensive datasets for action-object mapping.

Adaptability and Transferability: TBBF excels in generalization, allowing our system to handle new tasks and scenarios with minimal input—only a single image and one trajectory demonstration are required for online testing. By identifying therblig indices associated with specific actions, TBBF can adapt trajectories for new layouts by transferring task-related object changes across different environments. This adaptability is essential for real-world applications where tasks often vary dynamically.

As shown in Table 1, we benchmark TBBF against other architectures, demonstrating its superiority in task diversity, data efficiency, interpretability, and resilience in complex environments. This structured, therblig-based framework fundamentally improves robot learning and automation capabilities, addressing key limitations in existing end-to-end systems. The main contributions of this work include:

\begin{itemize} \item \textbf{Therblig-Based Backbone Framework (TBBF):} A structured framework that decomposes complex tasks into therbligs, enabling task decomposition, action-object mapping, and trajectory generalization. This framework enhances explainability, generalization, and efficiency in executing complex, long-horizon tasks. \item \textbf{Meta-RGate SynerFusion Network (MGSF):} A network for precise therblig segmentation, enhancing the understanding of sequential actions in robot tasks. \item \textbf{Action Registration (ActionREG):} A mechanism that integrates therbligs with object configurations, ensuring accurate action registration and stable task execution. \item \textbf{LLM-Alignment Policy for Visual Correction (LAP-VC):} A method that leverages large language models for visual error correction, reducing dependency on highly accurate demonstrations and enhancing adaptability. \end{itemize}

\section{Related research}

Various intelligent robot systems achieve high accuracy in specific tasks, such as cable routing \cite{luo2024multi}, cloth manipulation \cite{chen2024trakdis}, and fruit grasping \cite{lee2024synthetic}. However, they often struggle to generalize across different tasks. Recently, efforts have been made to develop systems that can handle a variety of tasks \cite{wang2023mimicplay, coarse2fine, diffuser}. However, these systems typically require large datasets or simple scenarios for generalization and lack a clear backbone for task understanding.

\begin{figure}[t] 
\centering
\includegraphics[width=\linewidth]{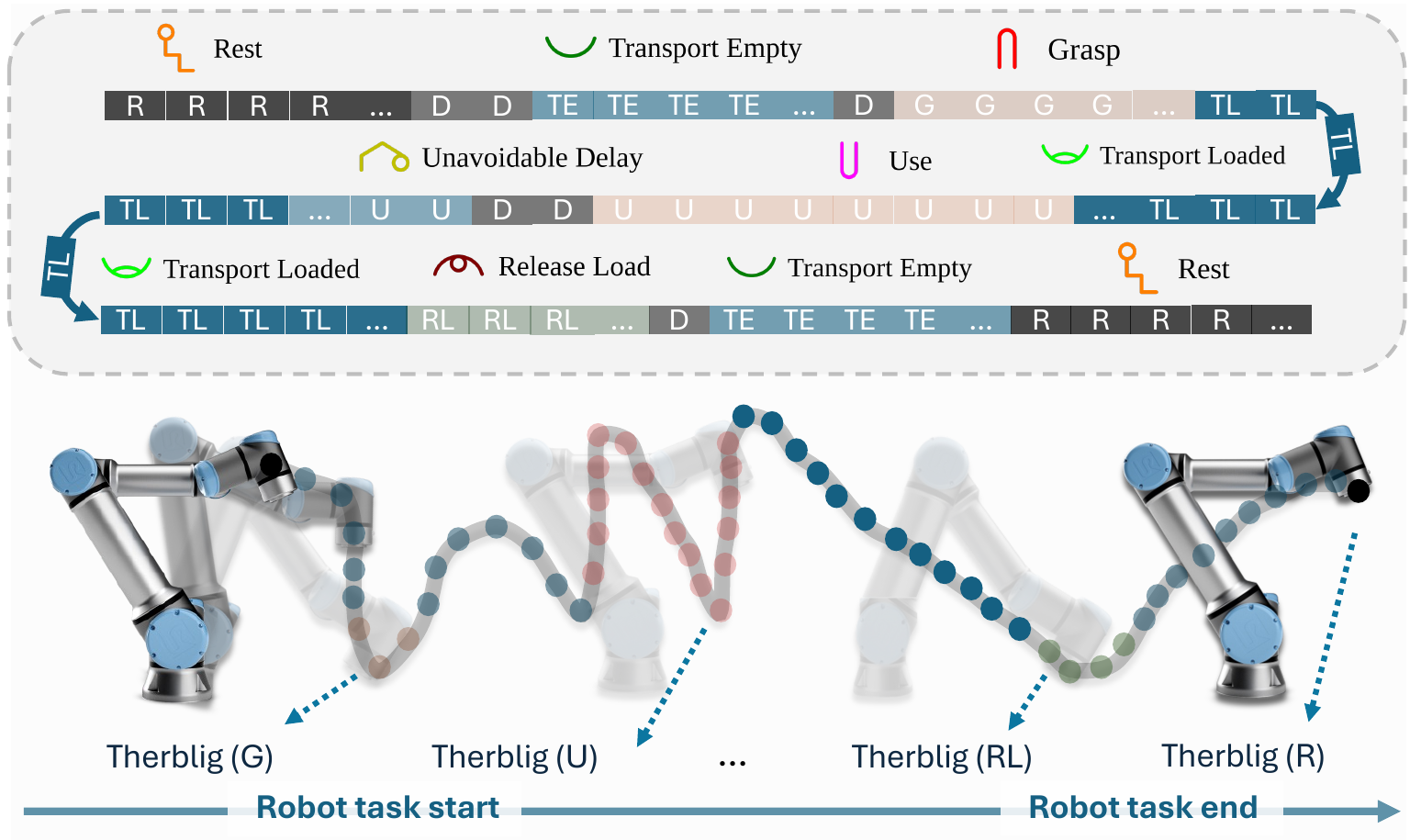} 
\caption{Detailed Decomposition of a Robotic Task into therbligs. The sequence containing: Rest (R), Transport Empty (TE), Delay (D), Grasp (G), Transport Load (TL), Use (U) and Release (R).}
\label{fig:actions}
\vspace{-3pt}
\end{figure}

\begin{figure*}[!ht]
\centering
\vspace{5pt}
\includegraphics[width=\textwidth]{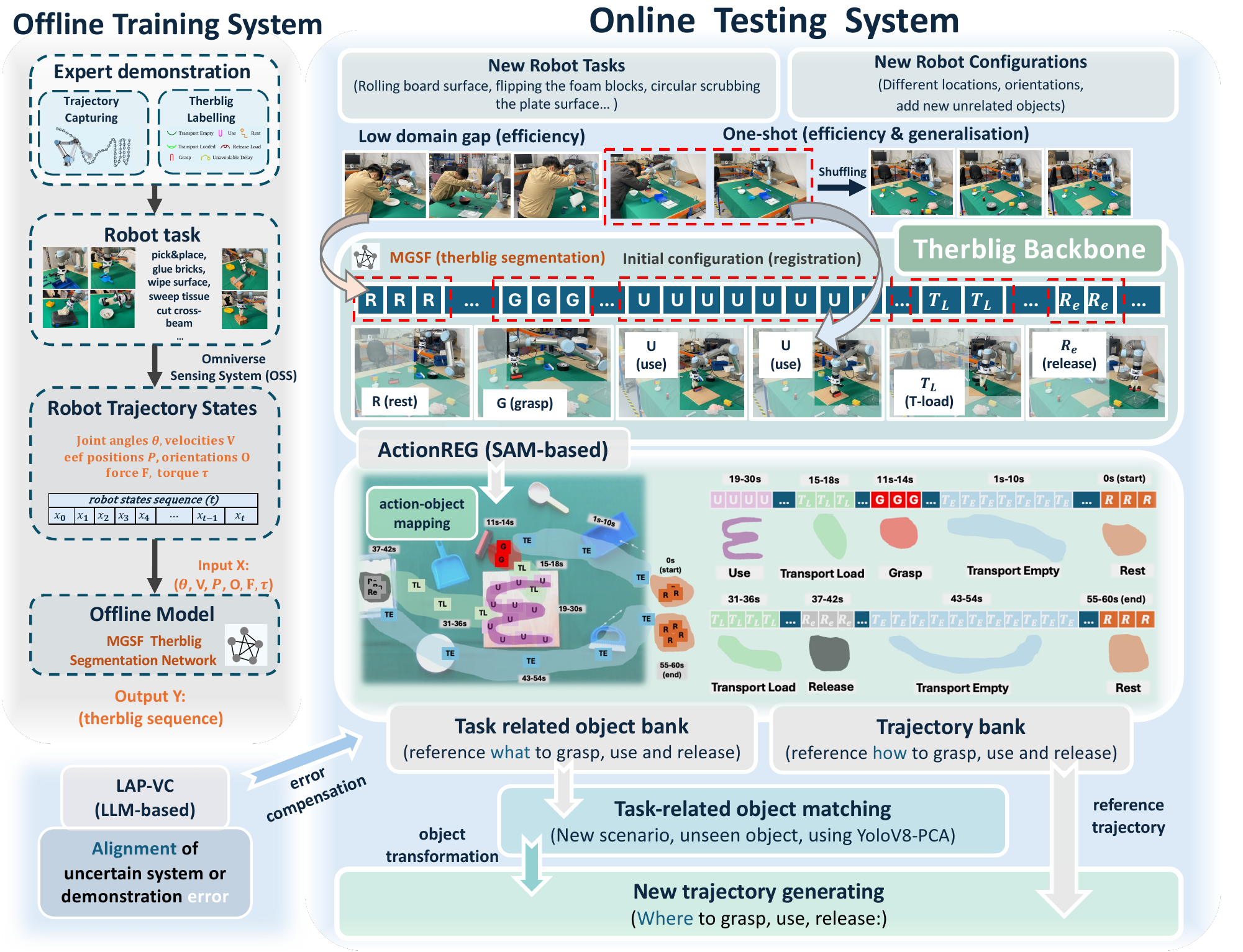}
\caption{Overview of the proposed \textit{TBBF}. This pipeline integrates offline training and online testing stages. During offline training, human experts provide demonstrations and label robot trajectories into therbligs, which are then used to train the \textit{MGSF} network. In the online testing stage, the trained \textit{MGSF} network segments new tasks into Therblig-level actions. \textit{ActionREG} registers these actions into new configurations, and \textit{LAP-VC} is utilized for error compensation. Finally, YOLOv8 and PCA are used to match new configurations. Arrows indicate the starting and ending points of the trajectory flow.}

\label{fig:pipeline}
\end{figure*}

\textbf{Kinematic and dynamic states-based Task Decomposition:} Ahmadzadeh et al. \cite{ahmadzadeh2015learning} introduced a method for converting action sequences into symbolic representations. Building on this, Chen et al. \cite{chen2020sequential} leveraged sequential motion primitives from human demonstrations, using a hierarchical BiLSTM classifier to extract intuitive high-level knowledge called therbligs. This approach enables more general representations for different task decompositions. However, their work remains conceptual, focusing on simple therblig segmentation without integrating vision modalities. This limitation results in reduced generalization and limited scene understanding.

\textbf{Video-based Task Decomposition:} For vision modalities, Dessalene et al. \cite{dessalene2023therbligs} introduced a rule-driven, compositional, and hierarchical action modeling method based on therbligs to analyze complex motions. This model features a novel hierarchical architecture comprising a Therblig model and an action model, utilizing vision as a medium for robot action segmentation. However, it lacks integration with action modalities based on human demonstrations, leading to a significant domain gap and inefficient capture of accurate task representations through images alone.

\textbf{Language-based Task Decomposition:} Large Language Models (LLMs) are utilized in robotics for task decomposition, each with advantages and limitations. Language instructions as input \cite{llm1wang2024grounding}\cite{llm2delta}\cite{llm3ahn2022can}\cite{huang2023voxposer}\cite{liang2023code} enable models like LLaMA and GPT-4 to quickly interpret high-level tasks and generate action sequences without detailed programming, but they lack precise trajectory data for executable paths. Moreover, systems that rely on manually designed action primitives often require extensive human engineering and can struggle to adapt to unseen or highly variable tasks, limiting their scalability. Such primitives may not capture the nuanced spatial or temporal details needed for complex, long-horizon tasks, and transferring them for new layouts.

\begin{figure*}[ht]
\centering
\vspace{5pt}
\includegraphics[width=\linewidth]{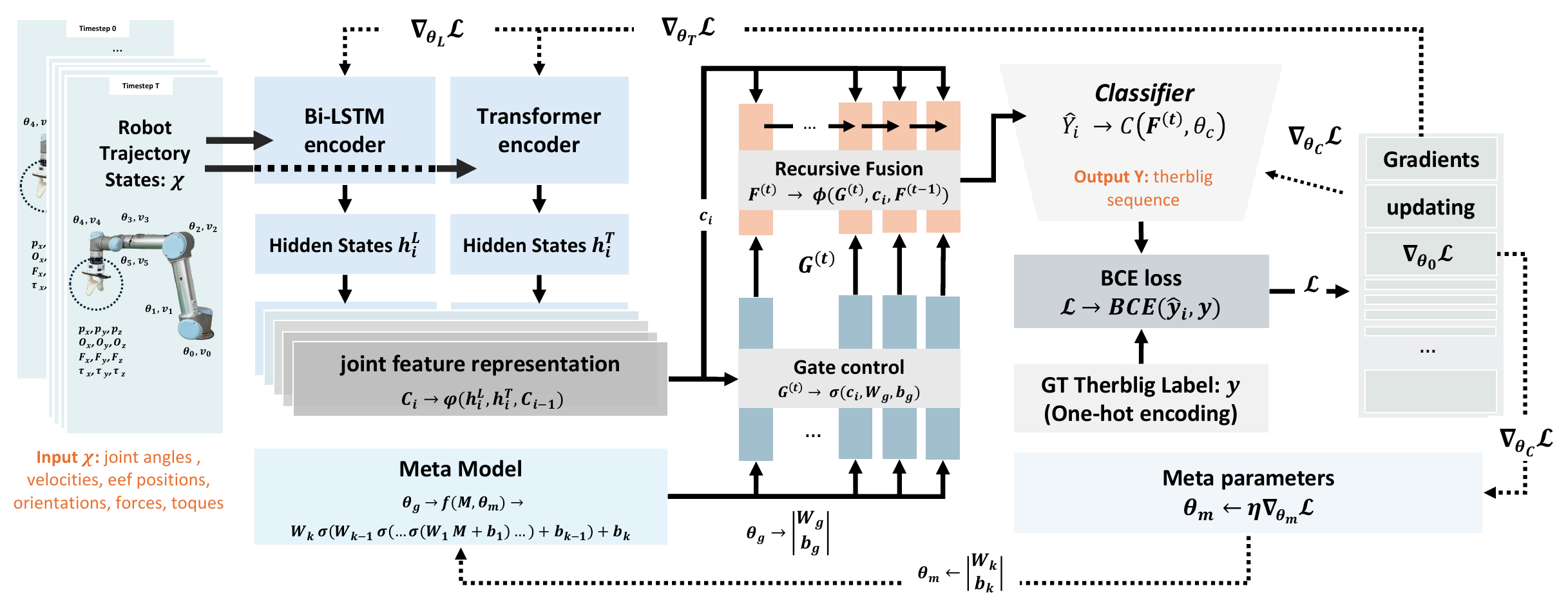} 
\caption{Detailed architecture of the \textit{MGSF} network. The \textit{MGSF} network integrates BiLSTM and Transformer sub-networks to capture sequential dependencies and use a meta-recursive gated fusion mechanism to dynamically combine the outputs of these sub-networks.}
\label{fig:neuralnetwork}
\end{figure*}

In our research, we propose to use therbligs as the backbone of a robot intelligent system to enhance task understanding. This TBBF represents a significant contribution towards a more structured, interpretable, and adaptable framework for robot task learning. By integrating this approach with the foundation model, we can easily extract the detailed configurations of the objects. Thus, we can create a more robust and flexible model for robotic systems.

\section{Model framework}
\subsection{\textit{TBBF}: Explainable robot task understanding framework}

The \textit{TBBF} is designed to enhance the understanding and generalization of robotic tasks by breaking them down into fundamental units called therbligs. This framework provides a structured and modular approach, facilitating better generalization across different tasks and scenarios while offering a clearer and more interpretable structure for task execution.

In the offline training stage, we utilize the \textit{MGSF} network to accurately segment tasks into therbligs, providing a detailed breakdown of the task into its constituent motions. During the online testing stage, we collect a one-shot demonstration of a new task, from which the \textit{MGSF} network extracts high-level knowledge and transforms it into a structured format. This knowledge is encoded into visual data using \textit{ActionREG}, integrating therbligs with the objects' configurations in the robot's visual field to ensure precise action registration. By using therbligs as the backbone, our framework significantly improves data efficiency and task generalization, enabling the robot to handle a wide range of scenarios with robustness and precision. The integration of \textit{LAP-VC} further ensures that any visual discrepancies are corrected in real time, providing an additional layer of accuracy in task execution. As depicted in Fig. \ref{fig:pipeline}, this advanced methodology enhances the robot's ability to adapt to new tasks by leveraging prior knowledge encoded in therbligs, thus improving  interpretability, stability, and transferability of robotic learning systems.

\begin{figure*}[t]
\vspace{5pt}
\centering

\includegraphics[width=\linewidth]{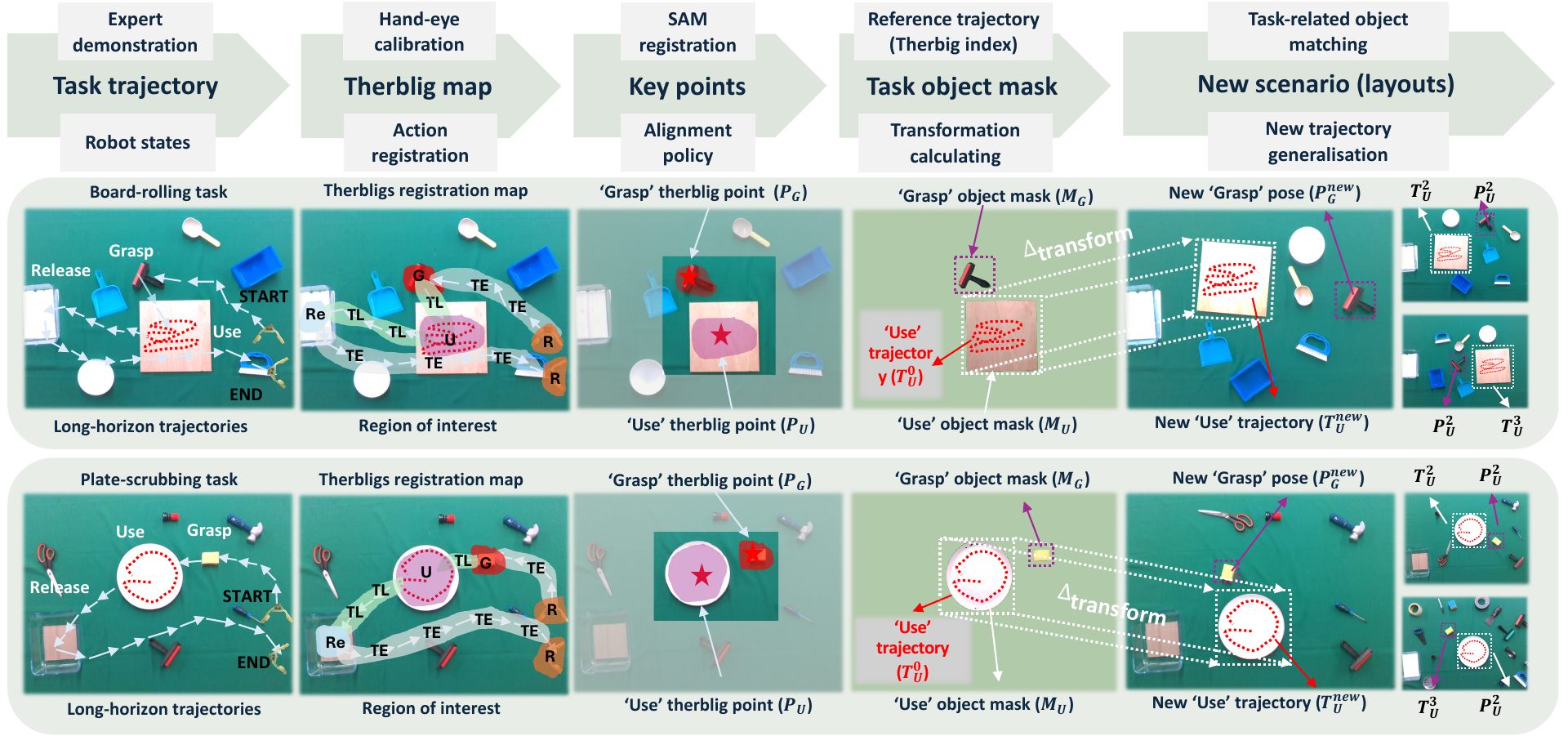} 
\caption{Details of the action registration, context matching, and new trajectory generating process. Arrows indicate the direction of trajectory.}
\label{fig:ActionReg}
\end{figure*}

\subsection{\textit{MGSF}: Efficient therbligs segmentation network}

\begin{algorithm}
\caption{\textit{MGSF} network for segmentation}
\label{alg:Meta-RGate SynerFusion}
\begin{algorithmic}
\small
\STATE \textbf{Input:} $X \in \mathbb{R}^{n \times d}$ \COMMENT{kinematic\&dynamic states sequence}
\STATE \textbf{Output:} $\hat{Y} \in \mathbb{R}^{n \times O}$ \COMMENT{therblig sequence}
\STATE Initialize: $\theta_L, \theta_T, \theta_g, \theta_m, \theta_c$
\STATE Initialize meta parameter vector $M \in \mathbb{R}^m$
\STATE Define the number of fusion steps $T \in \mathbb{N}$

\FOR{$i = 1$ to $n$}

    \STATE $h_i^L \gets \text{BiLSTM}(x_i; \theta_L)$ \COMMENT{BiLSTM hidden states}

    \STATE $h_i^T \gets \text{Transformer}(x_i; \theta_T)$ \COMMENT{Transformer hidden states}
    
    \STATE $c_i \gets [h_i^L \oplus h_i^T]$

    \STATE $F = F^{(0)}$ \COMMENT{Initial fusion output}

    \FOR{$t = 1$ to $T$}
        \STATE $\theta_g \gets f(M; \theta_m)$ \COMMENT{Gate parameters from meta-network}
        \STATE Compute gate values $G \in \mathbb{R}^{\dim(F)}$ using $\theta_g$
        \STATE $G = \sigma(\theta_g \cdot c_i)$ \COMMENT{Gate values using sigmoid function}
        \STATE Update fusion output $F \gets G \odot c_i + (1 - G) \odot F$
        \STATE $F = G \odot c_i + (1 - G) \odot F$ \COMMENT{Updated fusion output}
    \ENDFOR
    
    \STATE Compute predicted output $\hat{y}_i \gets \text{Classifier}(F; \theta_c)$
    \STATE $\hat{y}_i = \text{Classifier}(F; \theta_c)$ \COMMENT{Predicted output}
    \STATE Append predicted output to $\hat{Y} \gets \hat{Y} \cup \{\hat{y}_i\}$
    \STATE $\hat{Y} = \hat{Y} \cup \{\hat{y}_i\}$ \COMMENT{Update output set}
\ENDFOR
\end{algorithmic}
\end{algorithm}


\noindent \textbf{Algorithm 1 and Figure 4 Notation:} The input \( X \in \mathbb{R}^{n \times d} \) represents a sequence of task states (including joint angles, velocity, end-effector position, orientation, force, and torque), where \( n \) is the sequence length, and \( d \) is the feature dimension. For output, \( O \in \mathbb{R}^{n \times k} \) represents the one-hot encoded output, where \( n \) is the sequence length and \( k \) is the number of classes. Parameters \( \theta_L \) and \( \theta_T \) correspond to the BiLSTM and Transformer sub-networks, respectively, each designed to capture different aspects of the task’s sequential dependencies. The meta-parameter vector \( M \in \mathbb{R}^m \) is initialized to control the gated fusion mechanism dynamically, allowing the model to adapt to task-specific demands. Gate parameters \( \theta_g \) are generated from \( M \) and are used to dynamically control the fusion of BiLSTM and Transformer features, balancing between short-term and long-term dependencies. The recursive fusion step uses \( G^{(t)} \) at each time step \( t \) to combine features \( c_i \) and the previous fusion output \( F^{(t-1)} \) into the fused feature \( F \). The classifier layer, with parameters \( \theta_c \), uses the final fused feature \( F \) to output a predicted sequence of task labels \( \hat{Y} = [\hat{y}_1, \hat{y}_2, \dots, \hat{y}_n] \), where each \( \hat{y}_i \) corresponds to a predicted label for each time step in the sequence. The ground truth sequence \( Y = [y_1, y_2, \dots, y_n] \) represents the actual therblig labels (basic action elements) for each time step, with each label encoded in one-hot format. The Binary Cross-Entropy (BCE) loss \( \mathcal{L} \) is computed between the predicted sequence \( \hat{Y} \) and the ground truth sequence \( Y \) to guide the learning process. The Meta Model dynamically updates gate control parameters \( W_g \) and bias \( b_g \) for each task via meta parameters \( \theta_m \) to enhance adaptability across tasks.

Our \textit{MGSF} network is illustrated in Fig. \ref{fig:neuralnetwork} and detailed in Algorithm \ref{alg:Meta-RGate SynerFusion}. By combining meta-learning with adaptive gated fusion within a unified framework, this model significantly enhances robots' ability to comprehend and execute sequential actions across various environments. Inspired by MetaGross \cite{tay2020metagross}, our \textit{MGSF} network incorporates meta-gating and recursive parameterization in a recurrent model. However, MetaGross lacks a dedicated fusion process and struggles to integrate different aspects of the input data effectively, limiting its ability to leverage diverse features.

To address these limitations, our \textit{MGSF} model introduces a dynamic hybrid architecture that combines the strengths of both BiLSTM and Transformer sub-networks with a novel adaptive gated fusion mechanism. This architecture features a meta-recursive gated fusion unit that dynamically adapts to integrate model outputs, thereby enhancing performance across diverse tasks. Unlike the static gating in MetaGross, our adaptive gated fusion mechanism allows for more flexible and responsive integration of sequential data, ensuring that long-term dependencies are effectively captured and processed. By leveraging the strengths of both BiLSTM and Transformer sub-networks, the \textit{MGSF} network excels in handling complex sequences with greater precision. The meta-learning component dynamically adjusts to the changing context of tasks, ensuring that the model remains accurate and applicable across different situations.

\subsection{\textit{ActionREG}: SAM-driven action registration network}
A cornerstone of our TBBF is the ActionREG, designed for reasoning context information and configuration (Fig. \ref{fig:actions}). Directly using SAM with geometric masks to predict object points can be unstable in cluttered environments without prior information. Instead, ActionREG integrates therbligs’ prior knowledge into the SAM model, enabling accurate reasoning about the objects involved in robotic tasks. This integration guides the model to better understand and predict object configurations within a task-specific context, ensuring reliable performance in complex scenarios.

\begin{algorithm}
\caption{\textit{ActionREG} and Trajectory Generalization}
\label{alg:ActionREGandMatching}
\begin{algorithmic}
\small
\REQUIRE Online demonstration $\mathbf{D}$, Therbligs segmentation model $M_{\text{MGSF}}$, Hand-eye calibration matrix $\mathbf{H}$, SAM model $M_{\text{SAM}}$, YOLOv8 model $M_{\text{YOLO}}$, Background prior $\mathbf{B}$, New environment image $\mathbf{I}_{\text{new}}$, Reference image $\mathbf{I}_{\text{ref}}$
\ENSURE New Trajectory based on new layouts $\mathbf{T}_{\text{new}}( \mathbf{I}_{\text{new}})$

\STATE $\mathbf{S}_{\text{therbligs}} = M_{\text{MGSF}}(\mathbf{D})$
\STATE $\mathbf{K} = \{\text{Rest}, \text{TEmpty}, \text{Delay}, \text{Grasp}, \text{Use}, \text{TLoad}, \text{Release}\}$
\STATE $\mathbf{M} = \bigcup_{k \in \mathbf{K}} M_{\text{SAM}}(\mathbf{H} \cdot \mathcal{G}(\mathbf{S}_{\text{therbligs}}, k), \mathbf{B})$
\STATE $\mathbf{P}_{\text{demo}}, \mathbf{O}_{\text{demo}} = \mathcal{P}_{\text{demo}}(\mathbf{S}_{\text{therbligs}}, \mathbf{D})$

\FOR{each $\mathbf{m}_k \in \mathbf{M}$}
    \STATE $\mathbf{Box}_k = M_{\text{YOLO}}(\mathbf{I}_{\text{new}}, \text{ComputeArea}(\mathbf{m}_k))$
    \STATE $\mathbf{F}_{\text{new}} = \text{SIFT}(\mathbf{Box}_k)$, $\mathbf{F}_{\text{ref}} = \text{SIFT}(\mathbf{I}_{\text{ref}})$
    \STATE $\mathbf{M}_{\text{match}} = \text{FLANN}(\mathbf{F}_{\text{new}}, \mathbf{F}_{\text{ref}})$
    \STATE $\mathbf{P}_{\text{new}, k} = \text{ComputePosition}(\mathbf{M}_{\text{match}})$
    \STATE $\mathbf{O}_{\text{new}, k} = \text{PCA}(\mathbf{M}_{\text{match}})$
\ENDFOR

\STATE $\mathbf{\Delta}_{\text{transform}} = \mathcal{T}(\mathbf{P}_{\text{new}}, \mathbf{P}_{\text{demo}})$
\STATE $\mathbf{T}_{\text{new}} = \mathcal{A}(\mathcal{F}_{\text{demo}}(\mathbf{S}_{\text{therbligs}}, \mathbf{D}), \mathbf{\Delta}_{\text{transform}})$
\RETURN $\mathbf{T}_{\text{new}}$

\end{algorithmic}
\end{algorithm}

Through \textit{ActionREG}, we efficiently extract task-related object masks and workspace configurations. The process starts with object mask segmentation via the Segment Anything Model (SAM), denoted as \( M_{\text{SAM}} \). YOLOv8, represented as \( M_{\text{YOLO}} \), then detects bounding boxes \( \mathbf{Box}_k \) for each object. Features are extracted using SIFT, denoted by \( \mathbf{F}_{\text{new}} \) for new images and \( \mathbf{F}_{\text{ref}} \) for reference images. Feature matching is performed by FLANN, denoted as \( \mathbf{M}_{\text{match}} \), and object orientations are determined using PCA, denoted by \( \mathbf{O}_{\text{new}, k} \). The transformation \( \mathbf{\Delta}_{\text{transform}} \) is calculated using the function \( \mathcal{T} \), which maps object positions from the demonstration (\( \mathbf{P}_{\text{demo}} \)) to the new environment (\( \mathbf{P}_{\text{new}} \)). The function \( \mathcal{F} \) extracts the demonstration trajectory, and the new trajectory \( \mathbf{T}_{\text{new}} \) is generated by applying the transformation \( \mathbf{\Delta}_{\text{transform}} \) through the function \( \mathcal{A} \). Then it can generalize to new configurations.

\subsection{\textit{LAP-VC}: LLM-Alignment Policy for Visual Correction}
\begin{figure}[t] 
\centering
\vspace{5pt}
\includegraphics[width=\linewidth]{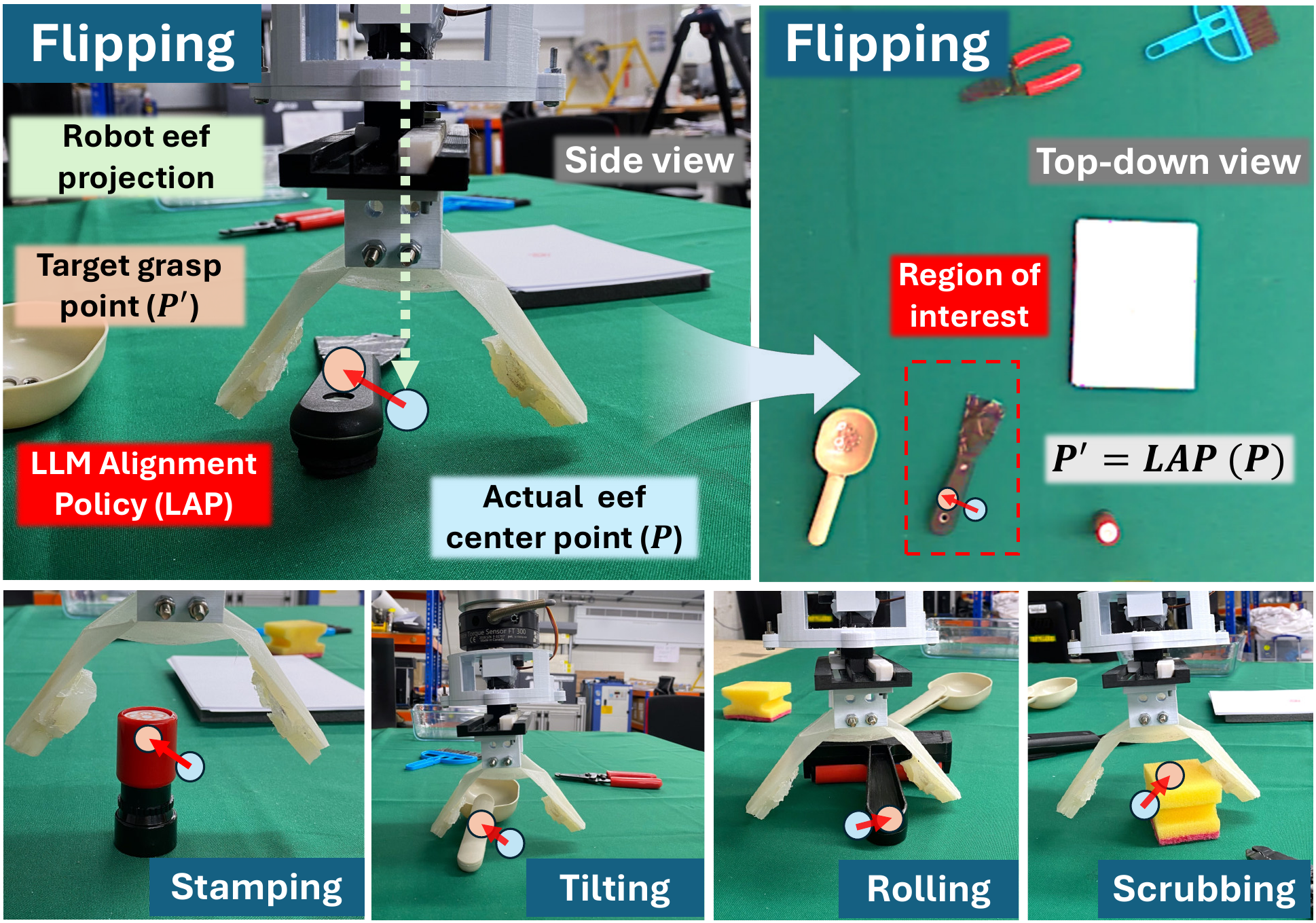} 
\caption{Application of \textit{LAP-VC} in robotic tasks. The pre-built prompt guides the LLM to process error points with scenario image and output corrected points, compensating for errors.}
\label{fig:LLM_align}
\vspace{-2pt}
\end{figure}

Expert demonstrations can have errors, such as the robot's end-effector not grasping perpendicularly, and hand-eye calibration inaccuracies. These issues can lead to incorrect position estimations, especially during the grasp stage, impacting action registration. To mitigate these challenges, we developed a novel method (Fig. \ref{fig:LLM_align}) that leverages LLM (GPT-4) for error correction. By feeding the predicted points and scenario image into the LLM with a pre-built prompt, the LLM provides corrected points. This approach minimizes the effects of imperfect demonstrations and system errors, reducing the reliance on highly accurate demonstrations.

\section{Experiments and Setup}

Offline data collection involved two individuals: one performing expert demonstrations and the other labeling the robot task's status. We gathered data from six tasks: tool pick-and-place, crossbeam cutting, bricks gluing, tissue sweeping, surface wiping, and cup pouring. These tasks were selected for their well-defined and repeatable patterns, covering a range of basic actions such as grasping, cutting, sweeping, wiping, and pouring. These tasks provided the model with diverse experiences to effectively learn essential behaviors. The offline training system utilized 52 groups of demonstrations per task, amounting to a total of 312 demonstrations. For each demonstration, we collected trajectory data over a 60-second timeframe (10 Hz), resulting in 600 timesteps. Each sample contained 26-dimensional data, including joint angles, joint speeds, end-effector poses and orientation, forces, and torques. Thus, the dataset for each demonstration was structured as a 600 × 26 matrix. The offline training set included 52 demonstrations (600 timesteps, 26 features each) and was split into 60\% training, 20\% validation, and 20\% testing. For online testing, a human expert performed new tasks in previously unseen scenarios. OSS recorded robot states and captured scenario images before and after each task. Five challenging tasks were used for testing: board rolling, foam block flipping, plate scrubbing, spoon tilting, and paper stamping. These tasks were selected to introduce novel movements, object interactions, and force dynamics which do not present in the training data.

\section{Results and Analysis}

For offline training, we used robot states to segment therbligs. Our \textit{MGSF} network outperforms state-of-the-art methods in this time-series segmentation task. As shown in \textit{Table \ref{tab:segmentation_performance}}, our network achieved an average recall of 94.37\% across 20 random seeds, surpassing methods such as TCNs, Reformer and the LLM-based BERT model. We also manually designed threshold-based methods; however, they were extremely time-consuming to configure, lacked generalizability across tasks, and delivered poor performance overall. In addition, we conducted an ablation study for \textit{MGSF} network to evaluate the impact of different components. The baseline recall of our backbone model (without fusion) was 79.77\% (transformer).  Introducing fusion mechanism increased the recall to 84.29\%, and adding a gate fusion mechanism further boosted it to 90.92\% (\textit{Table \ref{tab:segmentation_performance}}). Our dataset ablation study tested 30 to 312 demonstrations across six tasks. The model stabilized around 210-270 demos (35-45 per task), with no significant improvements at 312 demos, highlighting MGSF’s data efficiency and rapid convergence.

\renewcommand{\arraystretch}{0.7}

\begin{table*}[ht]
    \centering
    \vspace{5pt}
    \caption{Therblig Segmentation General Performance }
    \vspace{-2pt}
    \label{tab:segmentation_performance}
    \resizebox{\textwidth}{!}{  

    \begin{small}  
    \begin{tabular}{c c c c c c c}
        \specialrule{1.5pt}{1pt}{1pt} 

        \textbf{Benchmark Model} & \textbf{BCE-Loss $\downarrow$} & \textbf{Precision $\uparrow$} & \textbf{Recall $\uparrow$} & \textbf{F1-Score $\uparrow$} & \textbf{Kappa $\uparrow$} & \textbf{TP-Range $\uparrow$} \\
        \specialrule{1.5pt}{1pt}{1pt} 
        \textbf{TCNs}\cite{TCNs} & $0.623 \pm 0.002$ & $85.04 \pm 0.79$ & $88.48 \pm 0.79$ & $86.51 \pm 0.80$ & $85.54 \pm 0.99$ & $[85.45, 92.97]$\\
        \textbf{ABLG-CNN}\cite{deng2021attention} & $0.472 \pm 0.014$ & $25.32 \pm 0.55$ & $24.08 \pm 0.65$ & $21.18 \pm 0.98$ & $8.23 \pm 0.48$ & $[18.71, 32.05]$\\
        \textbf{MS-CRN}\cite{MS-CRN} & $0.058 \pm 0.008$ & $92.57 \pm 1.20$ & $92.55 \pm 1.19$ & $92.51 \pm 1.21$ & $90.76 \pm 1.48$ & $[75.58, 90.11]$\\
        \textbf{BiLSTM-T3}\cite{chen2020sequential} & $0.137 \pm 0.002$ & $78.84 \pm 0.56$ & $81.56 \pm 0.53$ & $79.74 \pm 0.51$ & $76.82 \pm 0.64$ & $[73.64, 86.56]$\\
        \textbf{GATv2}\cite{gatv2} & $0.145 \pm 0.008$ & $81.07 \pm 0.79$ & $82.22 \pm 0.63$ & $80.47 \pm 0.65$ & $77.58 \pm 0.78$ & $[91.88, 93.80]$ \\
        \textbf{Reformer}\cite{kitaev2020reformer} & $0.145 \pm 0.005$ & $77.87 \pm 0.69$ & $80.67 \pm 0.74$ & $78.79 \pm 0.72$ & $75.75 \pm 0.91$ & $[68.50, 88.17]$\\
        \textbf{TFT}\cite{tft} & $0.307 \pm 0.004$ & $69.37 \pm 0.46$ & $71.81 \pm 0.44$ & $69.28 \pm 0.41$ & $64.45 \pm 0.52$ & $[62.36, 82.02]$\\
        
        \textbf{TSMixer} \cite{tsmixer} & $0.125 \pm 0.005$ & $81.14 \pm 0.56$ & $83.51 \pm 0.44$ & $81.68 \pm 0.46$ & $79.25 \pm 0.56$ & $[80.06, 90.30]$ \\

        \textbf{LLM(Bert)} \cite{bert} & $0.092 \pm 0.017$ & $86.38 \pm 3.90$ & $86.89 \pm 2.96$ & $86.33 \pm 3.58$ & $83.67 \pm 3.78$ & $[85.92, 93.30]$ \\
        
        \specialrule{1.5pt}{1pt}{1pt} 
        \textbf{Ablation Model (descending)} & \textbf{BCE-Loss $\downarrow$} & \textbf{Precision $\uparrow$} & \textbf{Recall $\uparrow$} & \textbf{F1-Score $\uparrow$} & \textbf{Kappa $\uparrow$} & \textbf{TP-Range $\uparrow$} \\
        \specialrule{1.5pt}{1pt}{1pt} 

        \rowcolor{gray!30}\textbf{MGSF (Ours)} & $0.043 \pm 0.007$ & $94.36 \pm 0.60$ & $94.37 \pm 0.59$ & $94.36 \pm 0.60$ & $93.03 \pm 0.73$ & $[93.88, 97.17]$\\
        \textbf{GrNT (no meta)}\cite{GrNT} & $0.068 \pm 0.008$ & $90.96 \pm 0.83$ & $90.92 \pm 0.80$ & $90.84 \pm 0.83$ & $88.72 \pm 0.99$ & $[88.74, 94.73]$ \\ 
        \textbf{Adaptive-DF (no gate)}\cite{adaptive-FN} & $0.118 \pm 0.002$ & $81.86 \pm 0.29$ & $84.29 \pm 0.24$ & $82.44 \pm 0.25$ & $80.24 \pm 0.30$ & $[81.11, 89.82]$ \\
        \textbf{Backbone (no fusion)}\cite{vaswani2017attention} & $0.145 \pm 0.004$ & $77.00 \pm 0.92$ & $79.77 \pm 0.80$ & $77.94 \pm 0.82$ & $74.57 \pm 0.99$ & $[74.44, 87.67]$\\
        \specialrule{1.5pt}{1pt}{1pt} 
        \textbf{Ablation dataset} & \textbf{30 demos} & \textbf{90 demos} & \textbf{150 demos} & \textbf{210 demos} & \textbf{270 demos} & \textbf{312 demos (Ours)} \\
        \specialrule{1.5pt}{1pt}{1pt} 
        \textbf{MGSF (General Recall)} & $63.91 \pm 7.89$ & $76.06 \pm 3.45$ & $87.72 \pm 2.26$ & $92.46 \pm 0.99$ & $94.45 \pm 0.27$ & $94.37 \pm 0.59$\\
        \textbf{MGSF (General precision)} & $60.10 \pm 8.87$ & $78.01 \pm 2.72$ & $87.88 \pm 1.97$ & $92.47 \pm 1.01$ & $94.50 \pm 0.27$ & $94.36 \pm 0.60$\\
        \textbf{MGSF (F1-score)} & $59.15 \pm 11.37$ & $76.49 \pm 3.14$ & $87.70 \pm 2.20$ & $92.45 \pm 1.00$ & $94.46 \pm 0.28$ & $94.36 \pm 0.60$\\
        \specialrule{1.5pt}{1pt}{1pt} 
        
    \end{tabular}
    \end{small}
    }
\end{table*}

\renewcommand{\arraystretch}{0.9} 

\begin{table*}[ht]
    \centering
    \vspace{-5pt}
    \caption{Robot Task Success Rate Comparison (Long-horizon)}
    \vspace{-5pt}
    \small
    \label{tab:final_succesful_comparison}
    \resizebox{\textwidth}{!}{  
    \begin{tabular}{ccccccc}
        \specialrule{1.5pt}{3pt}{2pt} 
        \textbf{Model / Task} & \textbf{Board-rolling} & \textbf{FoamBlock-flipping} & \textbf{Plate-scrubbing} & \textbf{Spoon-tilting} & \textbf{Paper-stamping} & \textbf{Total} \\
        \specialrule{1.5pt}{1pt}{1pt} 

        SM + ST + SimScenario & 13/50 (26\%) & 2/50 (4\%) & 11/50 (22\%) & 0/50 (0\%)  & 15/50 (30\%) & 13.7\% \\

        BC + ST + SimScenario + one shot & 0/50 (0\%) & 0/50 (0\%) & 0/50 (0\%)  & 0/50 (0\%)  & 0/50 (0\%) & 0\% \\
        
        BC + ST + SimScenario + 100 shots & 6/50 (12\%) & 0/50 (0\%) & 0/50 (0\%)  & 2/50 (4\%)  & 0/50 (0\%) & 2.7\% \\

        BC + MT + SimScenario + one shot & 0/50 (0\%) & 0/50 (0\%) & 0/50 (0\%)  & 0/50 (0\%)  & 0/50 (0\%) & 0\% \\

        BC + MT + SimScenario + 100 shots & 0/50 (0\%) & 0/50 (0\%) & 0/50 (0\%)  & 0/50 (0\%)  & 0/50 (0\%) & 0\% \\

        LLM(TBBF) + MT + SimScenario + one shot & 34/50 (68\%) & 28/50 (56\%) & 33/50 (66\%)  & 31/50 (62\%)  & 35/50 (70\%) & 64.4\% \\
        
        \rowcolor{gray!20}\textbf{MGSF(TBBF) + MT + SimScenario + one shot (ours)} & \textbf{48/50 (96\%)} & \textbf{47/50 (94\%)} & \textbf{46/50 (92\%)}  & \textbf{48/50 (96\%)}  & \textbf{47/50 (94\%)} & \textbf{94.4\%} \\
        \specialrule{1.3pt}{1pt}{1pt} 
        
        SM + ST + ComScenario  & 0/50 (0\%) & 0/50 (0\%) & 0/50 (0\%) & 0/50 (0\%)  & 0/50 (0\%) & 0\% \\

        BC + ST + ComScenario + one shot & 0/50 (0\%) & 0/50 (0\%) & 0/50 (0\%) & 0/50 (0\%) & 0/50 (0\%) & 0\% \\
        
        BC + ST + ComScenario + 100 shots & 0/50 (0\%) & 0/50 (0\%) & 0/50 (0\%) & 0/50 (0\%) & 0/50 (0\%) & 0\% \\

        BC + MT + ComScenario + one shot & 0/50 (0\%) & 0/50 (0\%) & 0/50 (0\%) & 0/50 (0\%) & 0/50 (0\%) & 0\% \\
        
        BC + MT + ComScenario + 100 shots & 0/50 (0\%) & 0/50 (0\%) & 0/50 (0\%) & 0/50 (0\%) & 0/50 (0\%) & 0\% \\

        LLM(TBBF) + MT + ComScenario + one shot & 29/50 (58\%) & 24/50 (48\%) & 30/50 (60\%) & 26/50 (52\%) & 32/50 (64\%) & 56.4\% \\
        
        \rowcolor{gray!20}\textbf{MGSF(TBBF) + MT + ComScenario + one shot (ours)} & \textbf{43/50 (86\%)} & \textbf{42/50 (84\%)} & \textbf{40/50 (80\%)}  & \textbf{34/50 (68\%)}  & \textbf{41/50 (82\%)} & \textbf{80\%} \\
        
        \specialrule{2pt}{1pt}{1pt} 
        \textbf{Failure Case Analysis} & \textbf{Therblig Segmentation} & \textbf{Action Registration} & \textbf{Context Matching} & \textbf{Trajectory Planning} & \textbf{Others} & \textbf{Total} \\
        
        \specialrule{2pt}{1pt}{1pt} 
        
        Ours + SimScenario & 5/250 (2\%) & 2/250 (0.8\%) & 4/250 (1.6\%)  & 2/250 (0.8\%)  & 1/250 (0.4\%) & 5.6\% \\

        Ours + ComScenario & 8/250 (3.2\%) & 11/250 (4.4\%) & 9/250 (3.6\%) & 13/250 (5.2\%) & 9/250 (3.6\%) & 20\% \\
        
        \specialrule{1.3pt}{1pt}{1pt} 
        
    \end{tabular}
    
    }
    \noindent\begin{minipage}{\textwidth}
    \scriptsize
    \vspace{3pt}
    * SM means state machine methods, BC means behavior clone (shallow CNN based), LLM means large language model (BERT based), SimScenario means environment only contain task related objects (usually three objects), ComScenario mean environment contain many irrelevant objects (usually seven to ten objects), one shot means only provide one demonstration data (one image, one kinematic trajectory), 100 shots means providing input 100-demonstration data as input. Others in the failure study means factors such as  demonstration failure, LAP-VC failure, potential collision, system crash.
    \vspace{-7pt}
    \end{minipage}
\end{table*}

\begin{figure}[t] 
\centering
\vspace{5pt}
\includegraphics[width=\linewidth]{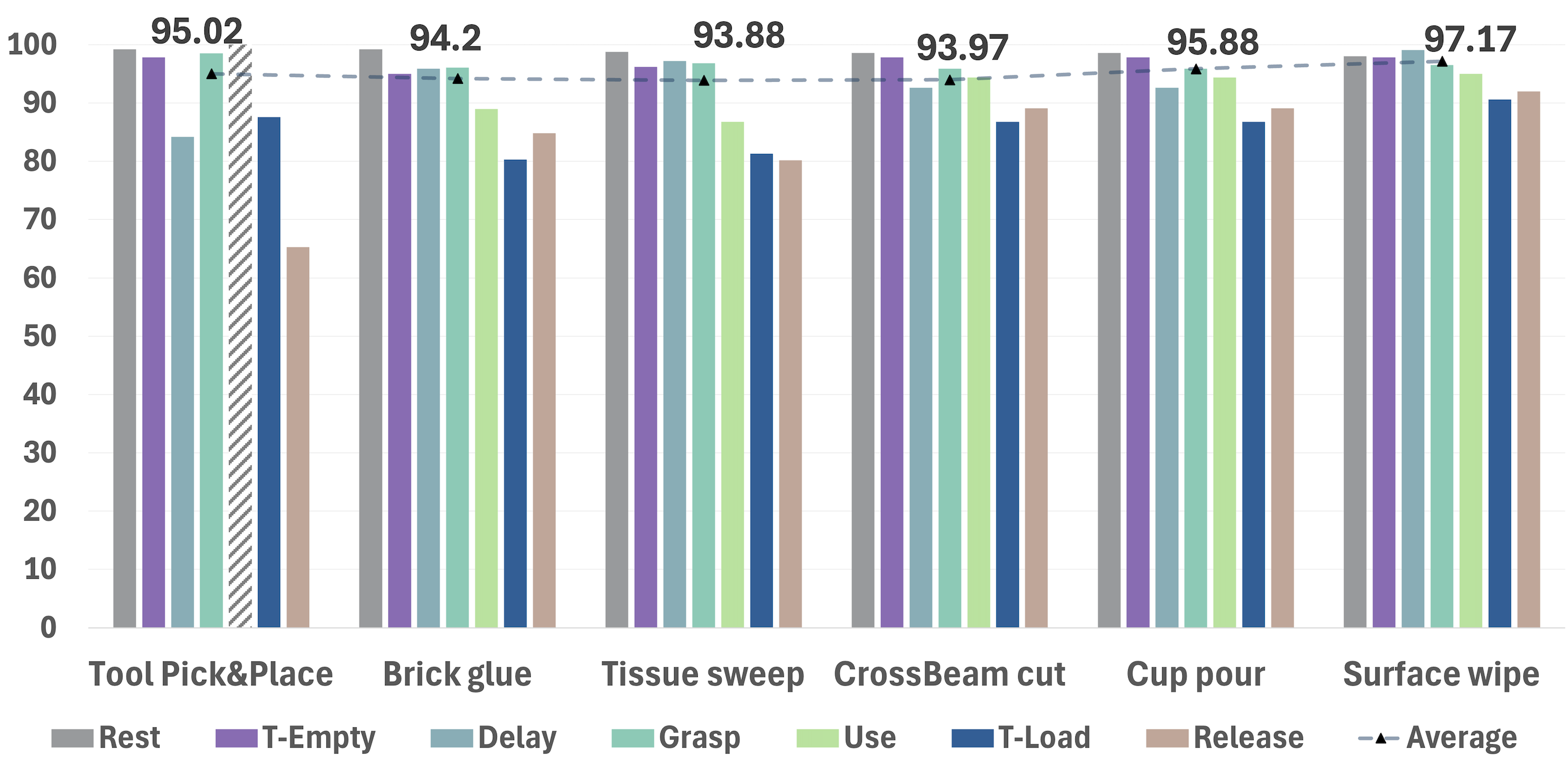}
\caption{Therblig segmentation recall for different robot tasks. For pick and place task, data points represented with an underscore indicate that the data is none with use. }
\label{fig:Fig7_Therblig_acc}
\vspace{-3pt}
\end{figure}

Moreover, we also analyzed the recall of different therbligs in terms of various tasks (Fig. \ref{fig:Fig7_Therblig_acc}). Note that these results are based on a single random seed and may slightly differ from the general recall results. Generally, surface wiping achieved the highest segmentation results (97.17\%) with six diverse robot tasks, while tissue-sweeping achieved the lowest results (93.88\%). This discrepancy may be attributed to the complexity of tissue-sweeping actions required to effectively put tissue into the dustpan. TLoad and Release achieved the lowest recall results, around 85.56\% and 83.41\% respectively. The lower release recall, caused by force-torque sensor drift, can occur after intricate manipulation operations.

\begin{figure}[t] 
\centering
\includegraphics[width=\linewidth]{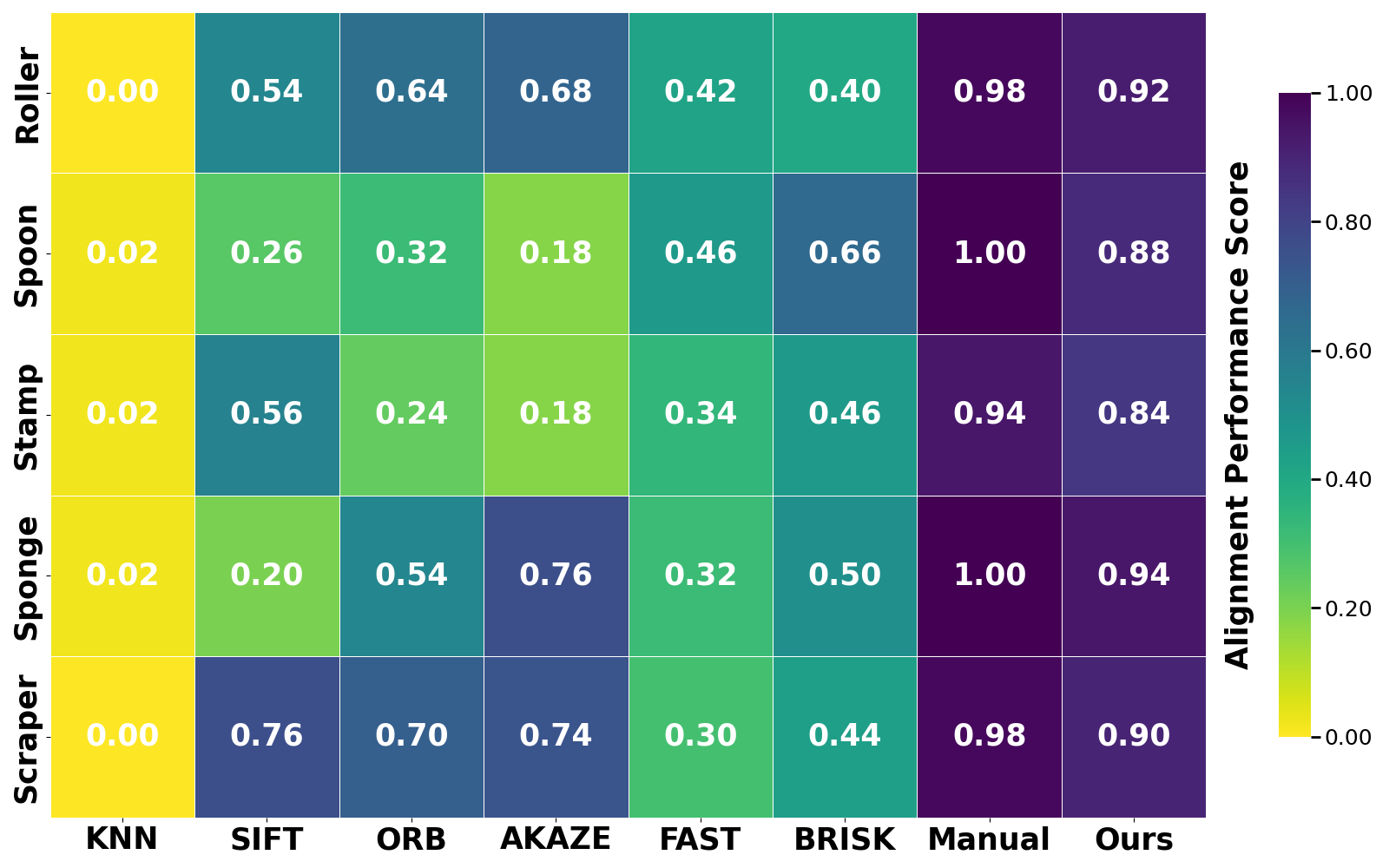}
\caption{Alignment performance comparison across various methods.}
\label{fig:Fig8_LLM_Benchmark}
\vspace{-12pt}
\end{figure}

Furthermore, the LAP-VC system (Fig. \ref{fig:Fig8_LLM_Benchmark}) consistently achieves high alignment performance scores, averaging 0.896 across various tasks, outperforming traditional methods such as KNN, SIFT, ORB, AKAZE, FAST, and BRISK. Its performance is slightly lower than that of human experts manually performing alignment. Specifically, for the Stamp task, the LAP-VC system achieved a score of 0.84. This lower score is attributed to the relatively small size of the stamp. In contrast, the Sponge task achieved a higher score of 0.94 due to sponge's relatively simple and uniform structure.

For the results of task execution, \textit{Table \ref{tab:final_succesful_comparison}} showed that our system can achieve promising performance for simple scenario (SimScenario) and complex scenario (ComScenario) with multi tasks and one shot data. For the SimScenario, we only consider two task-related and unseen objects appearing in image. For ComScenario, we add 3-6 unrelated and unseen objects together with our task-related objects to mimic the real-world cluttered environments. For SimScenario mode, our system achieved around 94.4\% average success rate for five tasks, while our success rate decreased to 80\% if we switched to ComScenario mode. We find that spoon-tilting has a lower success rate because it requires a more dynamic trajectory while robot solvers are easily trapped by singularity. We compared our system with State Machine, Behavior Clone and an LLM-based baseline. State machines, designed for single tasks, achieved some success in SimScenario with well-designed policies but struggled with increased complexity across tasks. For Behavior Cloning, both one-shot and 100-shot training were tested. Only single task and SimScenario with 100-shot works but most produced unstable trajectories prone to failure. The LLM baseline (with BERT) outperformed others when integrated into our one-shot TBBF but lagged behind our system due to lower therblig segmentation accuracy, demonstrating the critical impact of therblig segmentation module on task execution.

The failure case analysis underscores the interpretability of our TBBF system by identifying issues in specific modules. In SimScenario, the system demonstrates robustness with a low failure rate of 5.6\%, mainly due to therblig segmentation (2\%) and context matching (1.6\%). Even in ComScenario, our system maintains good performance with a failure rate of 20\%, where trajectory planning (5.2\%) and action registration (4.4\%) are the primary areas for improvement. This indicates that while our system performs reliably in both structured and complex environments, the ability to pinpoint module-specific issues allows us to enhance performance.

\section{Conclusion and Future Work}
We presented a novel Therblig-Based Backbone Framework (TBBF) that enhances the understanding and execution of robotic tasks by decomposing complex tasks into fundamental therbligs, which serves as the core architecture enabling all key modules in our system. This therblig backbone allows for improved data efficiency, interpretability, and generalization by providing a structured representation upon which our modules operate. Our experimental results demonstrate the effectiveness of our framework, showcasing high recall in therblig segmentation and robust performance in real-world robot task execution. We achieved results with 94.37\% recall in therblig segmentation and impressive successful execution rates of 94.4\% for new and long-horizon tasks in simple scenarios, and 80\% in complex scenarios. 

However, some limitations should be addressed in future work. Firstly, our offline training dataset is relatively small. We plan to use more efficient data collection methods to build a larger and more diverse dataset. Additionally, we focused on 2D object configurations and plan to extend our approach to 3D configurations while incorporating geometric constraints to prevent collisions in complex environments. We also plan to conduct a more detailed analysis of the therblig backbone in future work, focusing on how each component impacts the overall success rate and comparing its performance with manually designed action primitives. Furthermore, we plan to deploy a local LLM model, such as LLaMa3, to reduce latency and improve processing efficiency.

\vspace{+0.2cm}
\bibliographystyle{IEEEtran}
\bibliography{my_reference}

\end{document}